\documentclass{SCIS2021}

\begin{document}
\ArticleType{NEWS \& VIEWS}
\Year{2021}
\Month{}
\Vol{}
\No{}
\DOI{}
\ArtNo{}
\ReceiveDate{}
\ReviseDate{} 
\AcceptDate{}
\OnlineDate{}

\title{Comprehensive Benchmark Datasets for Amharic Scene Text Detection and Recognition}{Comprehensive Benchmark Datasets for Amharic Scene Text Detection and Recognition}

\author[1]{Wondimu DIKUBAB}{}
\author[1]{Dingkang Liang}{}
\author[1]{Minghui Liao}{}
\author[1]{Xiang BAI}{xbai@hust.edu.cn}

\AuthorMark{Xiang Bai}

\AuthorCitation{Wondimu DIKUBAB, Dingkang Liang, Minghui Liao, Xiang BAI}


\address[1]{1Huazhong University of Science and Technology, Wuhan {\rm 1037}, CHINA}

\maketitle


\begin{multicols}{2}

\section{Background}
\label{sec:background}

Ethiopic/Amharic script is one of the oldest African writing systems, which serves at least 23 languages (e.g., Amharic, Tigrinya) in East Africa for more than 120 million people. 

The Amharic writing system, Abugida, has 282 syllables, 15 punctuation marks, and 20 numerals. The Amharic syllabic matrix is derived from 34 base graphemes/consonants by adding up to 12 appropriate diacritics or vocalic markers to the characters. Unlike Latin alphabets, each Amharic character constitutes conjugation of consonants and vowels as a single syllable. The syllables with a common consonant or vocalic markers are likely to be visually similar and challenge text recognition tasks. Moreover, visual complexity, poor image quality, and intermittent text appearance cause failures of Amharic scene text detection and recognition.  

Recently, detecting and recognizing Latin and Chinese characters in natural scenes have progressed tremendously. However, the discussion on Amharic scripts detection and recognition is insufficient mainly due to the lack of public datasets. Recently, Addis et al.~\cite{1} presented the first private dataset for Ethiopic/Amharic scene text recognition, which contains 2,500 text images and lacks robustness.

\section{The Proposed Datasets}

\subsection{Text Detection}

We construct Amharic scene text detection datasets: the Amharic Real-world scene Text (HUST-ART) and the Amharic SynthText (HUST-AST)to address the problems mentioned in Sec.~\ref{sec:background}.

\begin{figure*}[t]
    \centering
    \includegraphics[width=0.98\textwidth]{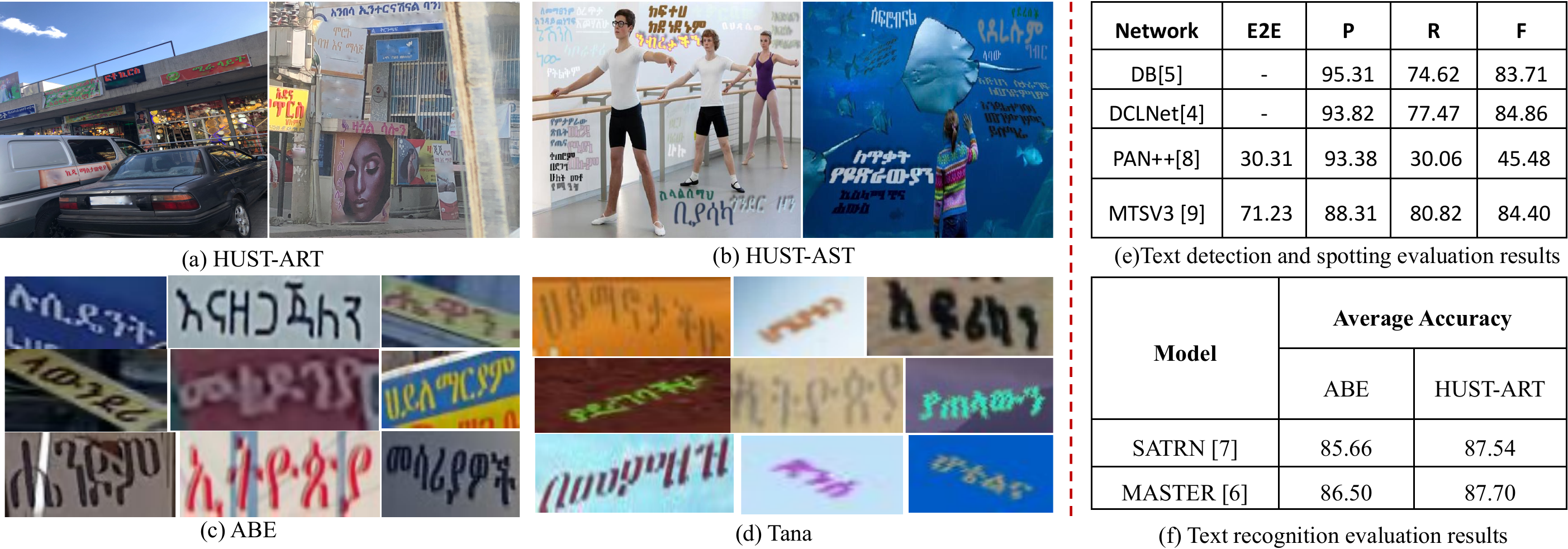}
    \vspace{-5pt}
    \caption{(a) Images from HUST-ART. (b) Images from HUST-AST. (c) Images from ABE. (d) Images from Tana. (e) Text detection and spotting results. (f) Text evaluation recognition results. E2E, P, R, and F refer to the End-to-End recognition rate, Precision, Recall, and F1-measure, respectively.}
    \label{fig:illustrate_dataset}
\end{figure*}

\textbf{HUST-ART} contains 2,200 natural scene images: 1,500 for the training and 700 for the testing. Specifically, it includes 11,254 cropped text instances. The HUST-ART pictures are collected across Ethiopia by mobile phone, professional cameras, and a few from the internet. This dataset comprises diversified scenes, including signboards, posters, indoors, streets, etc. We use quadrilateral coordinates to represent the ground truth of the text instance:
$G = [x_{1}, y_{1}, x_{2}, y_{2}, x_{3}, y_{3}, x_{4}, y_{4}]$, and word regions are categorized as easy or difficult.  The easy regions will be used for the recognition task (see Sec.~\ref{sec:recognition}). HUST-ART is robust and challenging in virtue of it contains multi-orientation text, small and large scale text, various illumination, and complex backgrounds, as shown in Fig.~\ref{fig:illustrate_dataset} (a). Moreover, HUST-ART has more text instances than the popular text detection dataset~\cite{2}.

\textbf{HUST-AST} contains 75,904 images with 829394 cropped synthetic text instances, and it is generated by SynthText~\cite{3} tool. The text sample is rendered upon natural images with random transformations and effects according to the local surface adaptation, as shown in Fig.~\ref{fig:illustrate_dataset} (b). 

\textbf{Evaluation.} We implemented SOTA methods DCLNet~\cite{4}, DB~\cite{5} to evaluate their performances on the proposed datasets.
Firstly, we use HUST-AST to pretrain the models, and then, we finetune the models on HUST-ART. Eventually, we select their final epoch for evaluation. As illustrated in Fig.~\ref{fig:illustrate_dataset} (e),we measure text detection performance by precision (P), recall (R), and F1-measure (F). DCLNet~\cite{4} achieves the best F1-measure of 84.67\%. Yet, we can see room for further improvement in the future.

\subsection{Text Recognition}
\label{sec:recognition}

Besides cropped word images from HUST-ART and HUST-AST datasets, we constructed two text recognition datasets of real-world and synthetic text, ABE and Tana, respectively.
\noindent
{\noindent\textbf{ABE} contains 12,839 real-word text images: 7,621 for training and 5,218 for testing.}
It is obtained by phone camera from Ethiopia and some from the Internet. The samples are shown in Fig.~\ref{fig:illustrate_dataset} (c). Compared with some previous datasets~\cite{1,2}, the proposed ABE contains more text images.

{\textbf{Tana} consists of 2,851,778 synthetic word images, including the 829394 HUST-AST cropped text images.} Besides HUST-AST, the text images are generated: applying random color, font rendering, blurring randomly, skewing the text arbitrarily, and blending with real-world images, as shown in Fig.~\ref{fig:illustrate_dataset} (d).

\textbf{Evaluation.} We adopt SOTA methods MASTER~\cite{6} and SATRN~\cite{7} to evaluate their Amharic scene text recognition performance on the proposed datasets ABE and HUST-ART. We use the Tana dataset as the training data, the union of ABE and HUST-ART training sets as validation data, and the ABE and HUST-ART testing sets as evaluation data. We measure the average accuracy rate by the success rate of word predictions per image. We only evaluate 302 character classes of syllables and Amharic numerals.

As the evaluation results in Fig.~\ref{fig:illustrate_dataset} (f) show, MASTER~\cite{6} outperforms both on ABE and HUST-ART datasets archiving 86.50\% and 87.70\%, respectively. The common causes of scene text recognition failure can be long text, blurred and distorted images, and uncommon fonts. Additionally, the Amharic scene text recognition failure can be caused by visual similarity among the characters that share a common consonant, the same kind of vocalic markers, or similar graphical structure. Therefore, the recognition of Amharic scripts requires more robust methods that can handle the visual similarity among the syllables.      

\subsection{End-To-End Text Spotting}

We train PAN++~\cite{9} and Mask TextSpotter v3 (MTSV3)~\cite{9} on joint HUST-AST and HUST-ART to evaluate their end to end text detection and recognition performance. We evaluate text spotting performance by precision(P), recall(R), F1-measure(F) and end-to-end recognition accuracy(E2E). The end-to-end text spotting performance evaluation results are presented in Fig.~\ref{fig:illustrate_dataset} (e). MTSV3 ~\cite{9} outperforms PAN++~\cite{8} achieving 71.23\% end-to-end recognition accuracy and 84.4\% F1-measure. 

Generally, the end-to-end text detection and recognition failure can be caused by inaccurate detection results, complex background with text-like patterns, the presence of irregular fancy text, low-resolution or blurred text, and false recognition results. Moreover, the evaluation results suggest that end-to-end Amharic text spotting demands more robust models.

\section{Conclusion}

In this work, we presented the first comprehensive public datasets named HUST-ART, HUST-AST, ABE, and Tana for Amharic script detection and recognition in the natural scene. We have also conducted extensive experiments to evaluate the performance of the state of art methods in detecting and recognizing Amharic scene text on our datasets. The evaluation results demonstrate the robustness of our datasets for benchmarking and its potential of promoting the development of robust Amharic script detection and recognition algorithms. Consequently, the outcome will benefit people in East Africa, including diplomats from several countries and international communities.

According to the quantitative results, we observed that the text detection and recognition performance demand a new attempt to design robust models that can address a unique feature of the Amharic script. We will dedicate ourselves to investigating the challenges and improving the detection and recognition performance in the future.

\lettersection{The datasets and more detailed information can be obtained from \url{https://dk-liang.github.io/HUST-ASTD/}}
\label{url:website}




%

\end{multicols}



\end{document}


\ArticleType{Supplementary File}

\title{Comprehensive Benchmark Datasets for Amharic Scene Text Detection and Recognition}{Comprehensive Benchmark Datasets for Amharic Scene Text Detection and Recognition}

\author[1]{Wondimu DIKUBAB}{}
\author[1]{Dingkang Liang}{}
\author[1]{Minghui Liao}{}
\author[1]{Xiang BAI}{xbai@hust.edu.cn}

\AuthorMark{Xiang Bai}
\AuthorCitation{Wondimu DIKUBAB, Dingkang Liang, Minghui Liao, Xiang BAI}


\address[1]{1Huazhong University of Science and Technology, Wuhan {\rm 1037}, CHINA}
\maketitle


\begin{appendix}

\section{Ethiopic/Amharic Writing System}
Amharic serves as an official working language of the Federal Democratic Republic of Ethiopia. It is the second-largest spoken Semitic language family next to Arabic globally. It is also used in Eritrea, Djibouti, Sudan, Somali Land, USA, Israel, Sweden as a business and second language. 

The Amharic/Ethiopic script is adapted from the Ethiopic syllabary, used for Geez, and developed in Ethiopia sometime during the 4th-century. The Ethiopic script has been adapted to write at least 20 different languages in Ethiopia, such as Tigrinya, Argobba, Awngi, Chaha, Harari, Sebat Bet, etc. It has conventionally been used for Tigrinya, Tigre, and Bilen in Eritrea. 

\begin{figure}[h]
 \centering
 \includegraphics[width=0.9\textwidth]{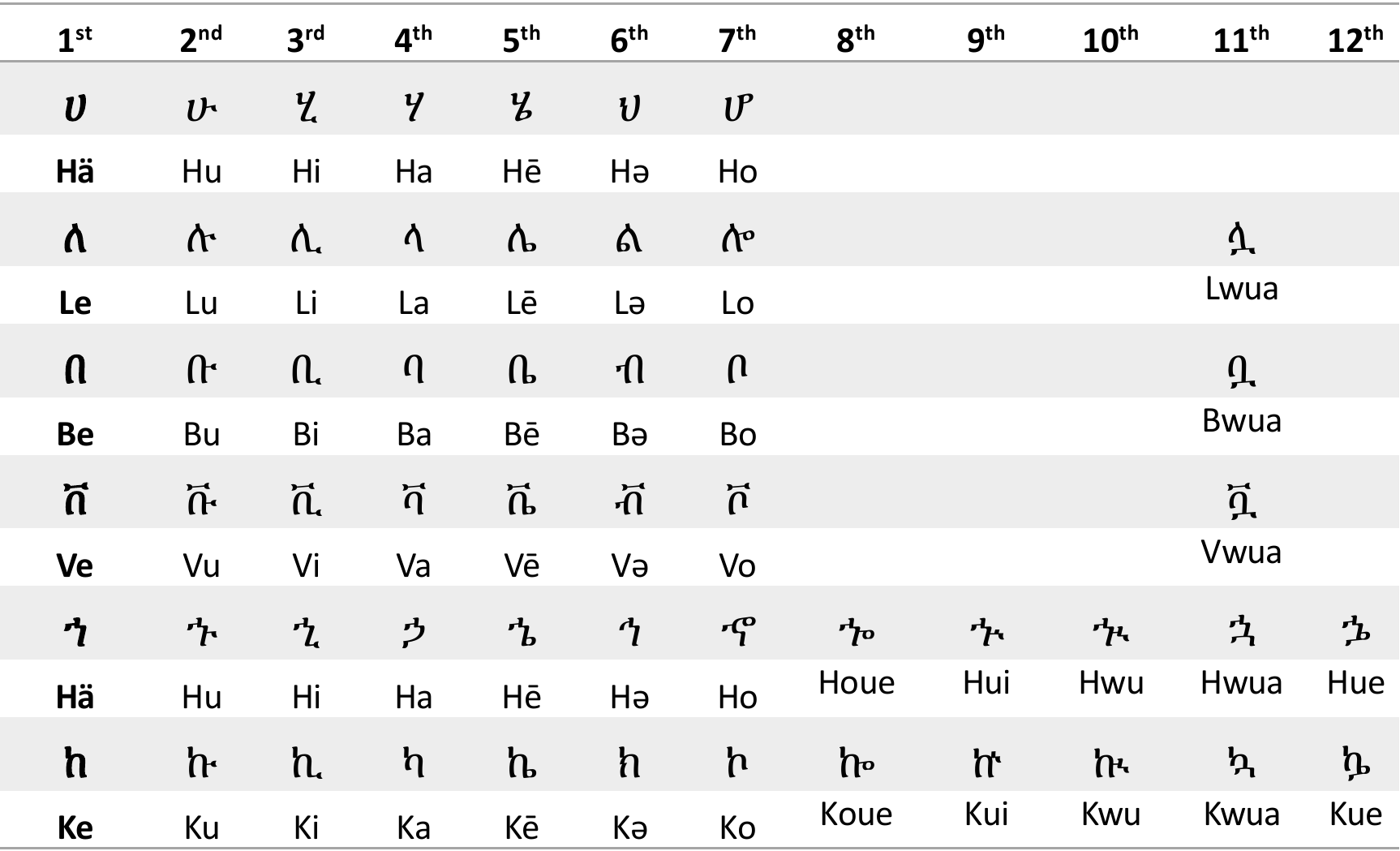}
 \caption{Example of Amharic Syllabary Matrix. All syllable in the same row inherits consonant sound and graphical shape from the first column. }
 \label{fig:Amharic_pro}
\end{figure}

The Amharic writing system is called Fidäl, Ethiopic or Abugida interchangeably. It has 282 syllables, 15 punctuation marks, and 20 numerals. The syllables of Abugida are derived from 34 base graphemes/consonants, which transformed into 248 syllabic symbols by adding appropriate diacritics or vocalic markers to the characters, as illustrated in Figure~\ref{fig:Amharic_pro}. 

The Ethiopic writing system is a featural syllabary, i.e., each Amharic character constitutes conjugation of consonants and vowels as a single syllable. The first 34 by seven Amharic Syllabary matrix is the core syllables. The others are known as Labiovelars and Labialized syllables. Labiovelar syllables (columns 8,9,10,12) are pronounced with the rounding of the lips, which are special Amharic characters. Labialized syllables (column 11) involve the lips while the remainder of the oral cavity produces consonant sound plus “wa” vocal. As illustrated in Figure~\ref{fig:Amharic_pro}, every Amharic character pronunciation represents a union of consonant and vowel sounds as an individual syllable. The pronunciation of each row is almost uniform, with few exceptions.
 
The Ethiopic writing system is univocal, and combining characters is not common. Unlike Latin, there is no upper and lower case distinction for Amharic characters. The Amharic script is written from left to right in horizontal lines.

\section{Text Detection }
We implemented SOTA methods such as DCLNet~\cite{1}, DB~\cite{2} and current popular methods, namely PSENET~\cite{3}, PAN~\cite{4} and EAST~\cite{5} to evaluate their performance on the proposed dataset.
Firstly, we use HUST-AST to pretrain the models, and then, we finetune the models on HUST-ART. Eventually, we select their final epoch for evaluation. As illustrated in Tab.~\ref{tab:det_result}, DCLNet~\cite{1} achieves the best F1-measure of 84.67\%.Yet, we can see room for further improvement in the future.

\begin{table}[h]
\footnotesize
\centering

\begin{tabular}{ |c|c|c|c|c|c| }
\hline
Method & Backbone & P (\%) & R (\%) & F (\%) & FPS \\ \hline
EAST~\cite{5}& Res50 & 79.67 & 79.10 &79.38 & 2  \\ \hline
PSENET~\cite{3}  & Res50 & 94.79     & 72.21  & 81.97   & 3.5   \\ \hline
PAN~\cite{4} & Res18 & 95.21     & 73.52  & 82.97   & 28  \\ \hline
DB~\cite{2}& Res18 & 96.61& 73.67 & 83.60 & 48  \\ \hline
DB~\cite{2} & Res50 & 95.31    & 74.62  & 83.71   & 22  \\ \hline
DCLNet~\cite{1} &Res50 & 93.82   & 77.47  & 84.86   & 3  \\ \hline
\end{tabular}
\caption{The detection performance results. P, R, and F refer to the Precision, Recall, and F1-measure, respectively.}
\label{tab:det_result}
\end{table}

\begin{figure}[h]
  \centering
  \includegraphics[width=\textwidth]{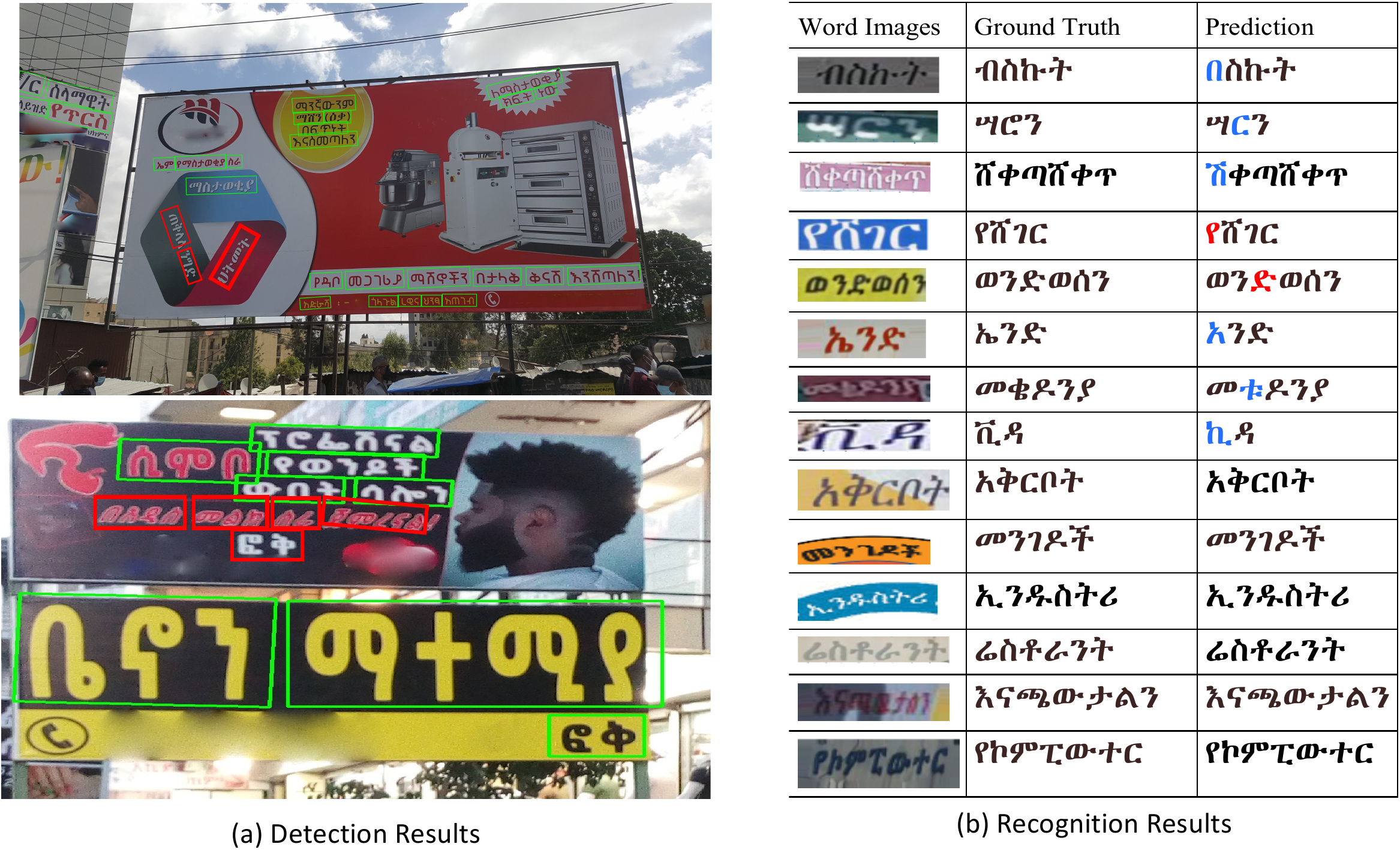}
  \caption{(a) Results of text detection. The green boxes are predictions, and the red boxes are either not predicted or miss predicted. (b) Results of recognition. The characters in blue color denote the wrong prediction, while the red color represents the characters missing.}
 \label{fig:det_recog}
\end{figure}

As we can observe from the qualitative evaluation samples in Figure~\ref{fig:det_recog} (a) the cause of the failure of the text detection can be text inside the text, low-resolution and small-sized text, text in rare rotation angle, etc.

\section{Text Recognition }

\begin{table}[h]
\centering
\footnotesize
\begin{tabular}{ |c|c|c| }
\hline
Method & ABE & HUST-ART   \\ 
\hline
CRNN~\cite{10} & 66.28\%  & 75.91\%  \\ \hline
RARE~\cite{9} & 72.08\%  & 80.46\%       \\ \hline
ASTER~\cite{8} &81.40\% & 85.30\% \\ \hline
SATRN~\cite{7} & 85.66\% & 87.54\% \\ \hline
MASTER~\cite{6} & 86.50\%  & 87.70\% \\ \hline
\end{tabular}
\caption{The recognition performance results. }
\label{tab:recognition_result}
\end{table}
\begin{figure}[h]
  \centering
  \includegraphics[width=\textwidth]{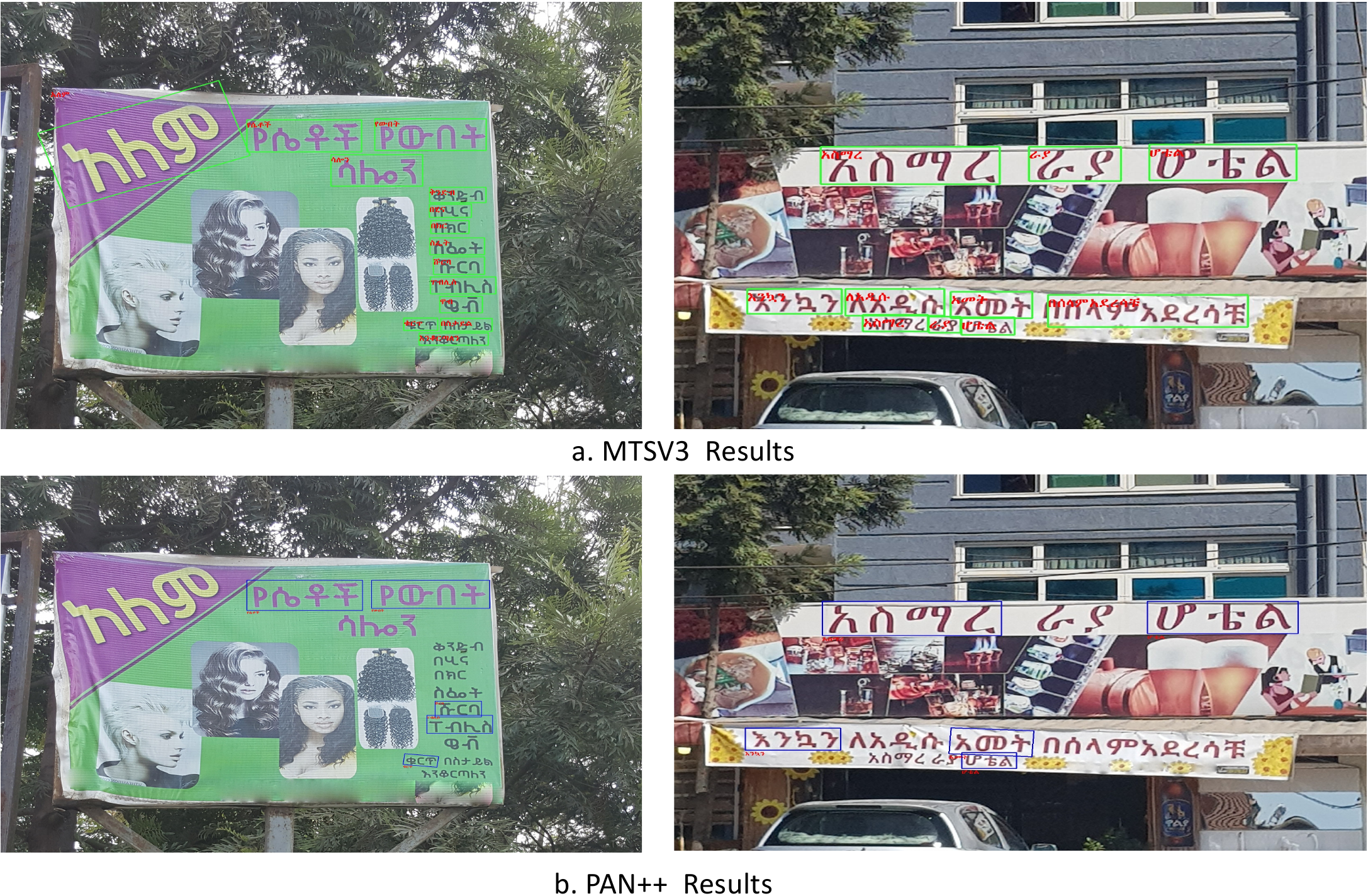}
  \caption{(End-to-end text spotting qualitative results.}
 \label{fig:spotter}
\end{figure}
We adopt SOTA methods such as MASTER~\cite{6} and SATRN~\cite{7}, ASTER~\cite{8} and current popular methods  RARE~\cite{9} and CRNN~\cite{10} to evaluate their Amharic scene text recognition performance on the proposed datasets ABE and HUST-ART. We use the Tana dataset as the training data, the union of ABE and HUST-ART training sets as validation data, and the ABE and HUST-ART testing sets as evaluation data. We measure the average accuracy rate by the success rate of word predictions per image. We only evaluate 302 character classes of syllables and Amharic numerals.

As the evaluation results in in Table.~\ref{tab:recognition_result} show, MASTER~\cite{6} outperforms both on ABE and HUST-ART datasets archiving 86.50\% and 87.70\%, respectively. 

As we can observe from the qualitative evaluation samples in Figure~\ref{fig:det_recog} (b)the common causes of scene text recognition failure can be long text, blurred and distorted images, and uncommon fonts. Additionally, the Amharic scene text recognition failure can be caused by visual similarity among the characters that share a common consonant, same vocalic markers, or similar graphical structures. Therefore, the recognition of Amharic scripts requires more robust methods that can handle the visual similarity among the syllables.

\begin{table}[h]
\footnotesize
\centering
\begin{tabular}{ |c|c|c|c|c| }
\hline
Method & E2E (\%) & P (\%) & R (\%)& F (\%) \\ \hline
PAN++~\cite{12} & 30.31   & 93.38 & 30.06 & 45.48   \\ \hline
MTPV3~\cite{11}& 71.23   & 88.31 & 80.82 & 84.40  \\ \hline
\end{tabular}
\caption{End to end text spotting quantitative results of the models on the HUST-ART dataset. E2E, P, R, F, and FPS refer to the End-to-End recognition rate, Precision, Recall, and F1-measure, respectively.}
\label{tab:spotting_result}
\end{table}

\begin{figure}[h]
  \centering
  \includegraphics[width=\textwidth]{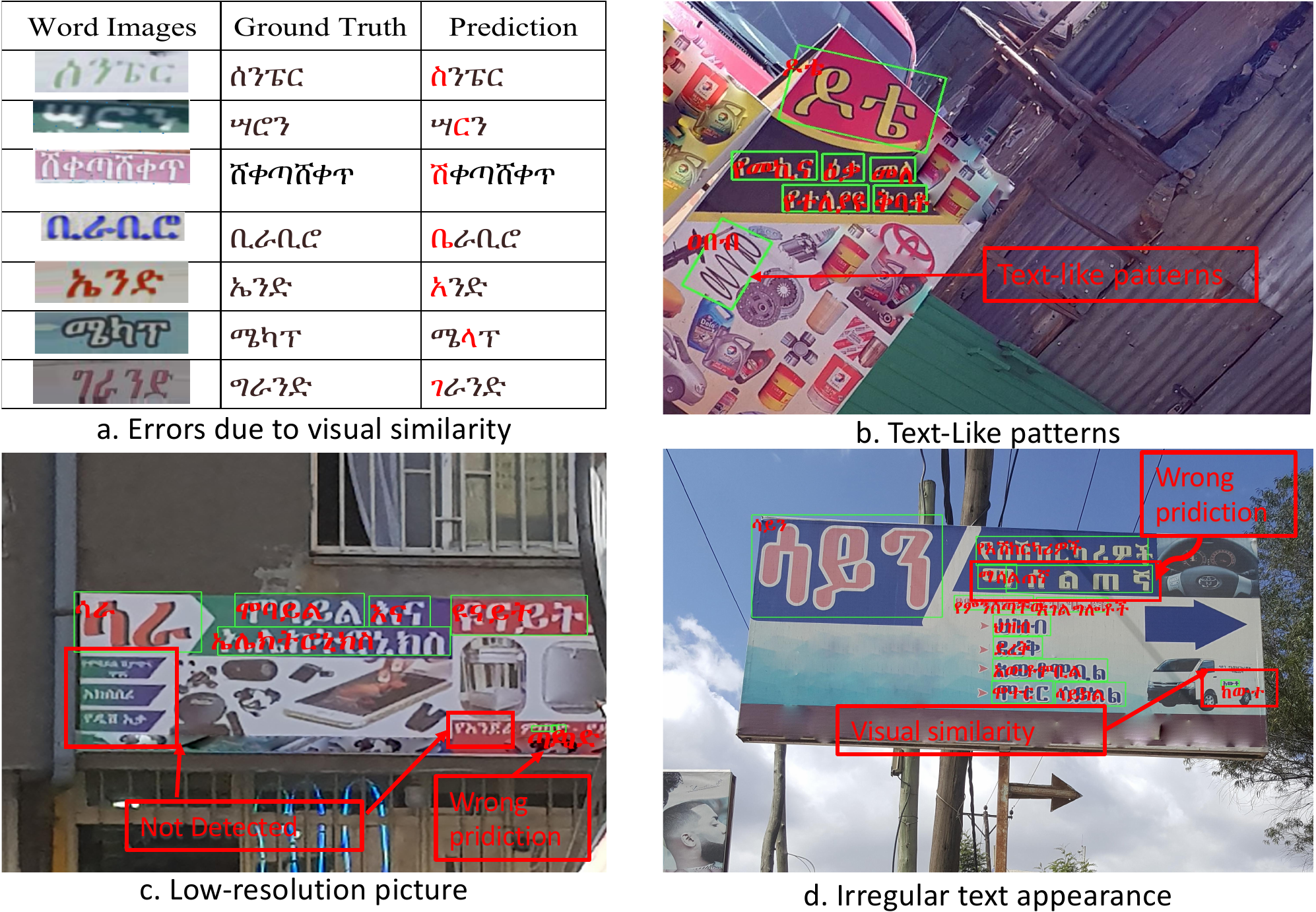}
  \caption{Failure cases: (a) Visual similarity, (b) Text-like patterns, (c) Low-resolution images, (d) Irregular text appearance. }
  \label{fig:Challange_images}
\end{figure}

\section{End-To-End Text Spotting}
We train Mask TextSpotter v3 (MTSV3)~\cite{11} and PAN~\cite{12} on joint HUST-AST and HUST-ART to evaluate their end-to-end text detection and recognition performance. 
We evaluate text spotting performance by precision(P), recall(R), F1-measure(F) and end-to-end recognition accuracy(E2E). The end-to-end text spotting performance evaluation results are presented in Tab.~\ref{tab:spotting_result}. MTSV3~\cite{11} outperforms PAN++~\cite{12} achieving 71.23\% end-to-end recognition accuracy and 84.4\% F1-measure. Generally, the end-to-end text detection and recognition failure can be caused by inaccurate detection results, complex background with text-like patterns, the presence of irregular fancy text, low-resolution or blurred text, and false recognition results. Moreover, the evaluation results suggest that end-to-end Amharic text spotting demands more robust models.

The qualitative end-to-end detection and recognition results in Figure~\ref{fig:spotter} show that MTSV3~\cite{11} performance is promising while PAN++[12] performance is insufficient. We now add these results in the revised paper.

\section{Challenges of Amharic Text Detection and Recognition}

We investigate the principal causes of the limitation of models to detect and recognize Amharic text in the wild. The challenges can be caused by visual similarity of characters, complex background, poor image quality, and style alignment.
\begin{itemize}
\item[(1)] The visual similarity among the characters is a unique nature of the Amharic writing system. The Amharic characters that share a common consonant, the same kind of vocalic markers, or similar structure are likely to be visually similar. Consequently, text recognition task becomes difficult not only for machines but also for humans (see Figure~\ref{fig:Challange_images} (a) ).
 
\item[(2)] Complex background scenarios with text-like patterns such as bricks, tree leaves, traffic signs, decorations, and fences, appear visually indistinguishable from the text 
(see Figure~\ref{fig:Challange_images} (b) ). The visual complexity causes errors and failures in scene text detection and recognition.

\item[(3)]The poor quality of the image due to the weather conditions and the intensity of illumination of the contextual scenery led to low-resolution, blurred, distorted, and skewed text images. Consequently, the low-quality pictures challenge scene text detection and recognition (See Figure~\ref{fig:Challange_images} (c) ).
\item[(4)] The irregular appearance of text in the wild comes with diverse font sizes, colors, multi-orientations, and text line patterns. Thus, the presence of irregular text cause failure to scene text detection and recognition algorithms (see Figure~\ref{fig:Challange_images} (d) ). 
\end{itemize}

\end{appendix}
